\documentclass[letterpaper, 10 pt, conference]{ieeeconf}  

\IEEEoverridecommandlockouts                              

\title{\LARGE \bf A Meta-Learning-based Trajectory Tracking Framework for UAVs under Degraded Conditions} 

\author{Esen Yel and Nicola Bezzo
	\thanks{Departments of Engineering Systems and Environment and Electrical and Computer Engineering, University of Virginia, Charlottesville, VA 22904, USA
		{\tt\small \{esenyel, nbezzo\}@virginia.edu}}%
}

\newcommand{\subparagraph}{}

\usepackage{xcolor}
\newcommand{\REV}[1]{{\color{black}#1}}

\usepackage{graphicx}
\usepackage{epstopdf}
\usepackage{amsmath}
\usepackage{amssymb}
\usepackage{subfigure}
\usepackage{multirow}
\usepackage{pbox}
\usepackage{algorithm}
\usepackage{algpseudocode}
\usepackage{titlesec}
\usepackage{bm}
\usepackage{dblfloatfix} 
\usepackage{float}
\usepackage{url}
 \usepackage{etoolbox}
\makeatletter
\patchcmd{\@makecaption}
  {\\}
  {.\ }
  {}
  {}
\makeatother

\DeclareMathOperator*{\argmin}{arg\,min}
 
\begin{document}

\maketitle
\thispagestyle{empty}
\pagestyle{empty}

\maketitle
\thispagestyle{empty}
\pagestyle{empty}

\begin{abstract}
Due to changes in model dynamics or unexpected disturbances, an autonomous robotic system may experience unforeseen challenges during real-world operations which may affect its safety and intended behavior: in particular actuator and system failures and external disturbances are among the most common causes of degraded mode of operation. To deal with this problem, in this work, we present a meta-learning-based approach to improve the trajectory tracking performance of an unmanned aerial vehicle (UAV) under actuator faults and disturbances  which have not been previously experienced. Our approach leverages meta-learning to train a model that is easily adaptable at runtime to make accurate predictions about the system's future state. A runtime monitoring and validation technique is proposed to decide when the system needs to adapt its model by considering a data pruning procedure for efficient learning. 
Finally, the reference trajectory is adapted based on future predictions by borrowing feedback control logic to make the system track the original and desired path without needing to access the system's controller.
The proposed framework is applied and validated in both simulations and experiments on a faulty UAV navigation case study demonstrating a drastic increase in tracking performance. 
\end{abstract}

\begin{section}{Introduction} \label{sec:intro}
	Autonomous mobile robots like aerial and ground vehicles are progressively becoming pervasive in our daily lives. Their widespread usage in various applications from transportation/delivery to surveillance increases their reliance accordingly. However, in real-world applications, robotic systems are subject to multiple challenges, such as external disturbances, model changes, and component failures. Because these situations usually occur without apriori knowledge, they are not considered during the design time; thus, they cause the system to operate under degraded conditions. 
	
	For example, let's consider an unmanned aerial vehicle (UAV) depicted in Fig.~\ref{fig:intro}, tasked to track a pipeline for inspection purposes. An unexpected failure or disturbance can compromise the stability of the system, making it deviate from the desired path possibly losing sight of the pipeline. Similarly, the same vehicle in a cluttered environment (e.g., heavy forested area) may end up deviating from a planned trajectory leading to possible unsafe situations like collisions.

		\begin{figure}[t]
		\centering
		\includegraphics[width=0.48\textwidth]{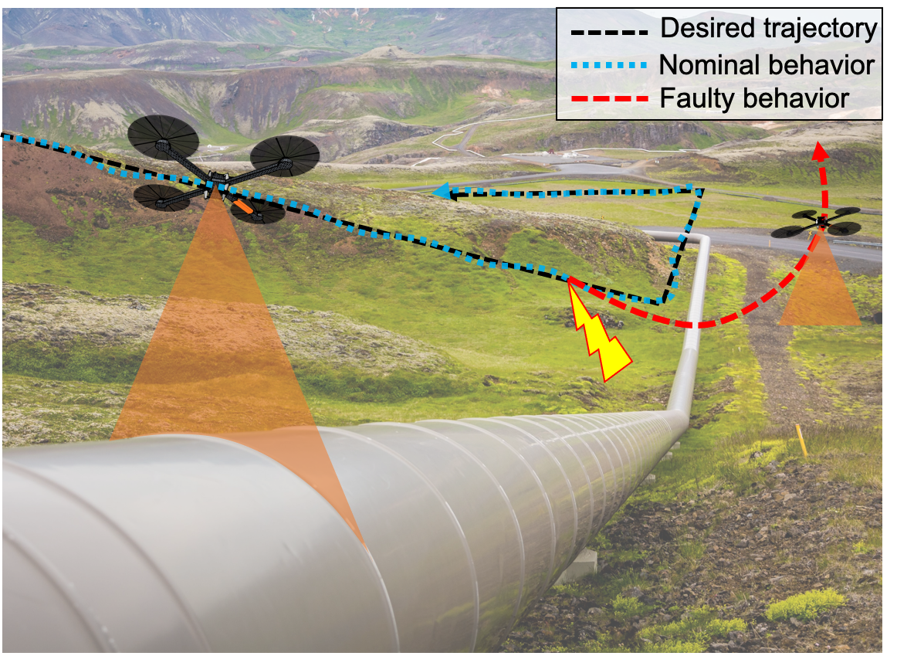}
		\vspace{-15pt}
		\caption{Pictorial representation of a UAV experiencing a failure at runtime causing task degradation like losing track of the pipe.}
		\label{fig:intro}
		\vspace{-20pt}
	\end{figure}

    To deal with such unforeseen and undesired situations, the system needs to quickly learn its model  under new conditions at runtime and adapt its behavior accordingly. To adapt the system's behavior, one can act on the system's controller, however, it is not always possible to reconfigure the system's controller, especially when off-the-shelf commercial vehicles with black-box control architectures are used. Instead, we observe that the high-level planner which is typically in charge of providing references for the existing controller can be easily adapted. 
    
    In order to update the reference trajectory, the system's model under new condition needs to be learned at runtime to make predictions about its future states. However, learning complex autonomous system models usually requires leveraging either 1) computationally demanding system identification and adaptive control techniques which may not be fast enough to characterize the new model online or 2) advanced machine learning techniques which may be limited by the amount and quality of training data. Among learning enabled components (LECs), deep neural networks (DNNs) have been demonstrated to be very effective to model complex system dynamics \cite{AbbeelICRA_15}. However, they typically require a large amount of data and long training time for accurate and reliable predictions. Meta-learning \cite{finn2017maml}, on the other hand, is a recent method developed to ``learn to learn" by leveraging optimization techniques to design machine learning models that can be adapted to the new tasks easily by using a few training examples. In this work, we propose a meta-learning based framework to predict the future states of an autonomous system under an unknown degraded condition and we use these predictions within our reference trajectory planning framework. Inspired by classical \REV{feedback} control theory we then propose to adapt the reference trajectory provided to the system to improve the trajectory tracking performance without accessing the controller. 

	The performance of the meta-learning system's model adaptation and prediction depends on the data collected at runtime. The initial data collected at runtime may not always be enough to represent the model as a whole. For this reason,  we introduce a runtime monitoring and validation framework to decide when the model needs to be re-learned due to inaccurate predictions. Additionally, to keep the runtime learning data size limited, we propose a data pruning approach for selecting the most representative data for the model.

	The contribution of this work is three-fold: 1) we develop a technique to update the reference trajectory of a UAV to improve its trajectory tracking performance; 2) we propose a runtime monitoring approach to decide when the model needs to be re-adapted, and 3) we propose a technique to select the data to quickly adapt the meta-learned model at runtime. 
	Even though the focus and motivation of this work is on faulty UAVs, the proposed approach is valid for improving the behavior of other types of robotic systems experiencing failures as well as disturbances.
	
	The rest of the paper is organized as follows: related work about failure rejection and meta-learning is presented in Section~\ref{sec:rel_lit}. The problem is then formally defined in Section~\ref{sec:problem} and we present our proposed meta-learning based trajectory tracking improvement approach in Section~\ref{sec:method}. We validate our technique with simulations and experiments in Sections~\ref{sec:simulation} and \ref{sec:experiments} respectively and finally draw conclusions and discuss future work in Section~\ref{sec:conclusion}.

\end{section}

\begin{section}{Related Work} \label{sec:rel_lit}
For consistent performance in challenging real-time conditions, control techniques have been widely used to adapt the systems' control inputs according to the changes in the system dynamics and to alleviate the effects of faults.
For quadrotors with the complete loss of one or multiple propellers, specific controllers can be designed according to the failure to improve stability and performance \cite{DandreaICRA2014, VisserRAL2018, indi_tro2020, HouAST2020}. These techniques require however the explicit knowledge of specific failures and how these changes affect the system's dynamical model in order to design resilient controllers. When such knowledge is not available, fault identification or adaptive control techniques need to be leveraged. For example, \REV{Model Reference Adaptive Control (MRAC) theory is combined with DNNs to learn the model uncertainty and applied for faulty systems in \cite{joshi2020asynchronous}.} In \cite{LeuteneggerRAL2020}, an Extended Kalman Filter (EKF)-based fault identification is used to decide if there is one or multiple rotor failures, and a control allocation is updated based on the failure using a nonlinear Model Predictive Control (MPC). In \cite{PhanRAL2020}, a self-reconfiguration technique which allows the system to decide on its configuration based on the actuator failure and its desired trajectory is introduced.

In addition to control approaches, machine learning techniques have been also widely used to improve the performance of UAVs under actuator failures or disturbances. In \cite{GombolayRAL2020}, the authors use MPC with active learning to learn the new model of the robot under failure and to provide necessary inputs. Reinforcement Learning (RL) techniques are also utilized to adjust the actuator control commands to compensate for component faults \cite{Fei2020LearntoRecoverRU, Caldwell2014}. 
Meta-learning approaches enable the systems to speed up their learning process for new tasks with small number of training samples from new tasks. Model Agnostic Meta-Learning (MAML) accomplishes this by training the model parameters explicitly to make them easy and fast to fine-tune for the new tasks \cite{finn2017maml}. MAML has been leveraged for fault-tolerant operations using MPC and RL \cite{nagabandi2019learning, ahmed2020complementary}. 
\REV{In \cite{Gombolaymeta2020}, a Q-function is learned via meta-learning and used to find optimal actions to maximize the probability of staying in the safe region for systems with altered dynamics. In \cite{pavone_rss_meta}, meta-learning is utilized to model the system dynamics under external forces to be used with an adaptive control scheme to improve the tracking performance.} 
All of these approaches assume that the user is given a direct access to the controller or the actuator inputs. However, this assumption may not hold, especially when off-the-shelf robotics systems are used. 

\REV{Machine learning techniques, such as DNNs, are utilized to design trajectories for good UAV trajectory tracking performance in \cite{Schoellig_impropmtu}.}
\cite{ZhouCDC2019} builds on this technique by improving the network training over time using active training trajectory generation based on the network prediction uncertainty. Even though these methods also propose trajectory update for improved tracking behavior, they are not designed for systems experiencing model changes between the design time and runtime; therefore, they are not suitable for systems experiencing component failures as presented in this paper. In this work, we also adapt the reference trajectory to improve trajectory tracking, but we consider systems under disturbances and faults. Our approach  utilizes MAML to \REV{train an adaptable network to predict the future states of a new faulty system. The predictions are then used to update the reference trajectory to minimize the deviation from the desired behavior. Different from the existing fault-tolerant approaches, our framework does not require the explicit knowledge of the system model, or access to the controller inputs. It improves the tracking performance of an existing controller by modifying the reference trajectory.}


\end{section}

\begin{section}{Problem Formulation} \label{sec:problem}
In this work, we are interested in finding a technique to predict the future states of a system under unforeseen actuator faults, to validate and relearn the prediction model at runtime, and to adapt system's reference trajectory to improve its trajectory tracking performance according to the predictions. These problems are formally defined as follows:

 \begin{figure*}[t]
    \vspace{2pt}
	\centering
	\includegraphics[width=0.98\textwidth]{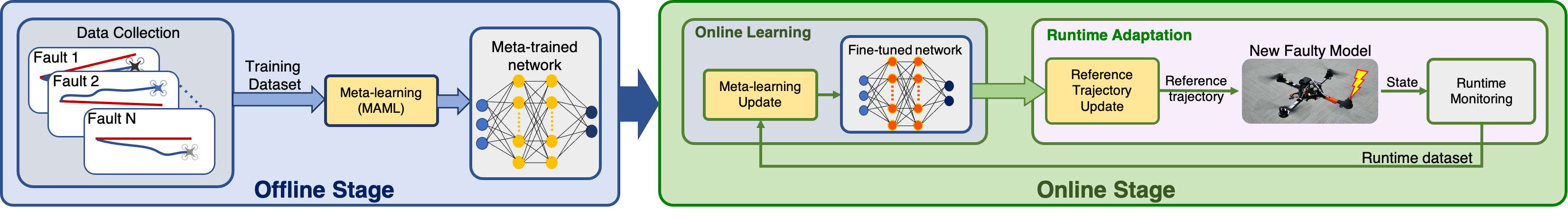}
	\vspace{-10pt}
	\caption{Proposed meta-learning-based framework for trajectory tracking recovery under unknown failures/disturbances.}
	\label{fig:diagram}
	\vspace{-13pt}
\end{figure*}	
	
\textbf{Problem 1: \textit{Future State Prediction under Failure}:}
A UAV with a nominal dynamical model $ f(\bm{x}, \bm{u}) $ as a function of its states $\bm{x}$ and controller inputs  $ \bm{u} $ has the objective of following a predefined desired trajectory $ \bm{x}_\tau $. Under actuator failures and disturbances 
the system's model changes to $\bm{x}(k+1) = f'(\bm{x}(k), \bm{u}(k))$.
In either case, control inputs are generated by a fixed controller $ \bm{u}(k) = g(\bm{x}(k), \bm{x}_\tau(k+1)) $. With these premises, find a policy to predict the future state of the system as a function of its current state and reference trajectory: $ \tilde{\bm{x}}(k+1) = \tilde{f}'(\bm{x}(k), \bm{x}_\tau(k+1)) $.


\REV{
\textbf{Problem 2: \textit{Runtime Monitoring:} }
As the prediction model can vary or change over time, the data which are used to update the model in the beginning of the operation may become unable to represent the model. To overcome this problem, design a runtime monitor to decide when the learned model becomes inaccurate, and to select the online learning data to effectively relearn the model at runtime.}

\textbf{Problem 3: \textit{Reference Trajectory Update}:}
\REV{Once the predictor for faulty system is learned and} the future state of the system is predicted by solving Problem 1, find an online policy to dynamically update the reference trajectory input  ($\tilde{\bm{x}}_\tau(k+1)$)  to the system such that the deviation from the original desired trajectory ($ d =  \rVert \bm{x} - \bm{x}_\tau \lVert $) is minimized.


\end{section}

\begin{section}{Trajectory Tracking Improvement Using Meta-Learning}  \label{sec:method}
	Our proposed framework consists of offline and online stages, as demonstrated in Fig.~\ref{fig:diagram}. 
	During the offline stage, UAVs facing various actuator faults are tasked to follow a set of trajectories at different speeds. For each actuator fault, the state of the system and the desired trajectory are recorded in the offline dataset, and a model for future position prediction is trained with this dataset using meta-learning. In this work, we use MAML \cite{finn2017maml} as the meta-learning approach due to its ability to train neural networks easy and fast to fine-tune with a small amount of data.
	
	At runtime, a UAV with a different failure than the ones used during the offline stage is tasked to follow a trajectory towards a goal. Once the vehicle starts its operation and collects a certain amount of data, \REV{the meta-trained model is fine-tuned according to the new observed data.} The fine-tuned neural network is then used to predict the system's future state according to its reference trajectory. Our trajectory update approach adaptively updates the reference trajectory according to the predicted deviation from the desired path. This approach allows us to trick the system into trying to follow a reference trajectory different than the actual desired trajectory; however, it, in fact, makes the UAV follow its original desired trajectory with a smaller deviation. With the proposed runtime monitoring approach, the meta-learned model is constantly validated and updated when necessary. In the next sections, we will explain each component of our approach more in details.

	\subsection{MAML for State Prediction under Degraded Conditions}
	In this work, during the offline stage, we create a dataset by collecting data from a UAV with an actuator fault/disturbance from a discrete fault set $ \mathcal{F} $ following different trajectories. This fault set also contains the case in which a UAV without a fault follows the same trajectories. At runtime, we aim to learn a model, denoted as $ \bar{f}' $ which takes as input the desired position and velocity with respect to the current position and velocity, and outputs the system's next position with respect to its current position.  
	A  training input $ \bm{x}^i_\text{DNN} \in \mathbb{R}^6 $ and a training output $ \bm{y}^i_\text{DNN} \in \mathbb{R}^3 $ for a UAV with the fault  $ \mathcal{F}_i \subset \mathcal{F} $ are calculated as follows:
	\begin{align}
		& \bm{x}^i_\text{DNN} =   \left[ \begin{array}{c}
		\bm{p}_{\tau_j}(k+1) \\ \bm{v}_{\tau_j}(k+1)		
		\end{array}\right]  -  \left[ \begin{array}{c}
		\bm{p}^i(k) \\ \bm{v}^i(k)	 	
		\end{array}\right]  \nonumber   \\
		& \bm{y}^i_\text{DNN} = \bm{p}^i(k+1) - \bm{p}^i(k) \label{eq:training} \\ 
		& \forall j \in {1, \ldots, N}, \forall k \in {1, \ldots, T(\tau_j)}  \nonumber
	\end{align}
	where $ N $ is the number of training trajectories, $ \bm{p}^i $ and $ \bm{v}^i $ are the position and velocity of the UAV under fault $ i $ respectively and $ T(\tau_j) $ is the duration of the trajectory $ \tau_j $ in unit time steps. The dataset $ \mathcal{D}_i $ for the fault $ \mathcal{F}_i $ contains the training input matrix $ \bm{X}_\text{DNN}^i \in \mathbb{R}^{6 \times M_i} $ and $ \bm{Y}_\text{DNN}^i \in \mathbb{R}^{3 \times M_i}$ with the columns $ \bm{x}^i_\text{DNN} $ and $ \bm{y}^i_\text{DNN} $ respectively. $ M_i $ is the number of data samples for fault $ \mathcal{F}_i $.	
	The training dataset for meta-learning contains the data from each failure: $ \mathcal{D}_i \in \mathcal{D} $. It should be noted that we used position and velocity as training inputs, however, our approach is independent of this choice and higher dimensional inputs can also be used depending on the complexity of the behavior and the prediction.

	The purpose of meta-learning is to train a model which is adaptable for 
	\REV{new}
	tasks at runtime. As a review, MAML considers a model represented by a parametrized function $ \bar{f}'_{\bm{\theta}} $ with parameters $ \bm{\theta} $. For ease of notation, for the rest of the paper, we will use $ f_{\bm{\theta}} $ to indicate the meta-learning model. During the offline meta-training, the model parameters $ \bm{\theta} $ are meta-optimized according to Algorithm 1 in \cite{finn2017maml}. In summary, $ \bm{\theta} $ is initialized randomly and updated to $ \bm{\theta}_{i}^{'} $ while adapting to fault $ \mathcal{F}_i $:
	\begin{equation}
	    \bm{\theta}_i^{'} = \bm{\theta} - \alpha \nabla_{\bm{\theta}} \mathcal{L}_{\mathcal{F}_{i}}(f_{\bm{\theta}})
	\end{equation}
	Meta-optimization across the tasks then updates the model parameters according to Equation 1 in \cite{finn2017maml}:
	\begin{equation}
	    \bm{\theta} \leftarrow \bm{\theta} - \beta \nabla_{\bm{\theta}} \sum_{\mathcal{F}_i \subset \mathcal{F}} \mathcal{L}_{\mathcal{F}_{i}}(f_{\bm{\theta}_{i}^{'}})
	\end{equation}
	where $ \alpha $ and $ \beta $ are hyperparameters for optimization step size and $ \mathcal{L} $ is the loss function. As our problem is a supervised learning problem, the loss function is given as follows:
	\begin{equation}
	    \mathcal{L}_{\mathcal{F}_{i}}(f_{\bm{\phi}}) = \sum_{\bm{x}^{i}_\text{DNN}, \bm{y}^{i}_\text{DNN} \in \mathcal{D}_i } \rVert f_{\bm{\phi}}(\bm{x}^{i}_\text{DNN}) - \bm{y}^{i}_\text{DNN} \lVert_2^2
	\end{equation}
	where $ \bm{x}^{i}_\text{DNN} $ and $ \bm{y}^{i}_\text{DNN} $ are given in \eqref{eq:training}.

	\subsection{Online Meta-Network Update} \label{sec:online_maml}
	At runtime,	as the UAV experiences a fault outside of the training fault set, 
	it collects data from its position and velocity sensors upon starting its operation and adapts its offline meta-learned model. 
	The system is able to quickly collect enough data to adapt its model as the model parameters are optimized for easy adaptation with meta-learning. 
	Initially, the vehicle collects $ K $ consecutive data and constructs an online learning dataset with input $ \bm{X}_\text{DNN}^* \in \mathbb{R}^{6\times K} $ and output $ \bm{Y}_\text{DNN}^* \in \mathbb{R}^{3\times K} $. The input and output sample data from the dataset are calculated as follows:
	\begin{align}
		& \bm{x}^{*}_\text{DNN}(k) = \left[ \begin{array}{c}
		\bm{p}^*_{\tau}(k+1) \\ \bm{v}^*_{\tau}(k+1)		
		\end{array}\right]  -  \left[ \begin{array}{c}
		\bm{p}^*(k) \\ \bm{v}^*(k)		
		\end{array}\right]  \nonumber \\
		& \bm{y}^{*}_\text{DNN}(k) = \bm{p}^*(k+1) - \bm{p}^*(k) 
	\end{align}	
	where $ k \in {1, \ldots, K} $, $ \bm{p}^* $ and $ \bm{v}^* $ are the position and velocity of the UAV under unknown fault $ \mathcal{F}^* $ respectively and $ \bm{p}^*_{\tau} $ and  $ \bm{v}^*_{\tau} $ are desired trajectory positions and velocities respectively. 
	\REV{With this runtime dataset, the meta-trained model parameters are updated with gradient descent updates
	and a fine-tuned model $ \bm{f}_{\theta^*} $ with updated parameters $ \bm{\theta}^* $ is obtained.} The fine-tuned model is then used and re-adapted to update the reference trajectory as explained in the following section.
	\subsection{Runtime Reference Update, Monitoring, and Re-learning}
	After the initial adaptation of the meta-trained model, the system constantly updates its reference trajectory to improve trajectory tracking and performs runtime monitoring and relearning for model validation by following the architecture demonstrated in Fig.~\ref{fig:diagram_runtime}. The reference trajectory is updated using the state predictions as explained next.
	\begin{figure}[b]
	    \centering
	    \includegraphics[width=0.48\textwidth]{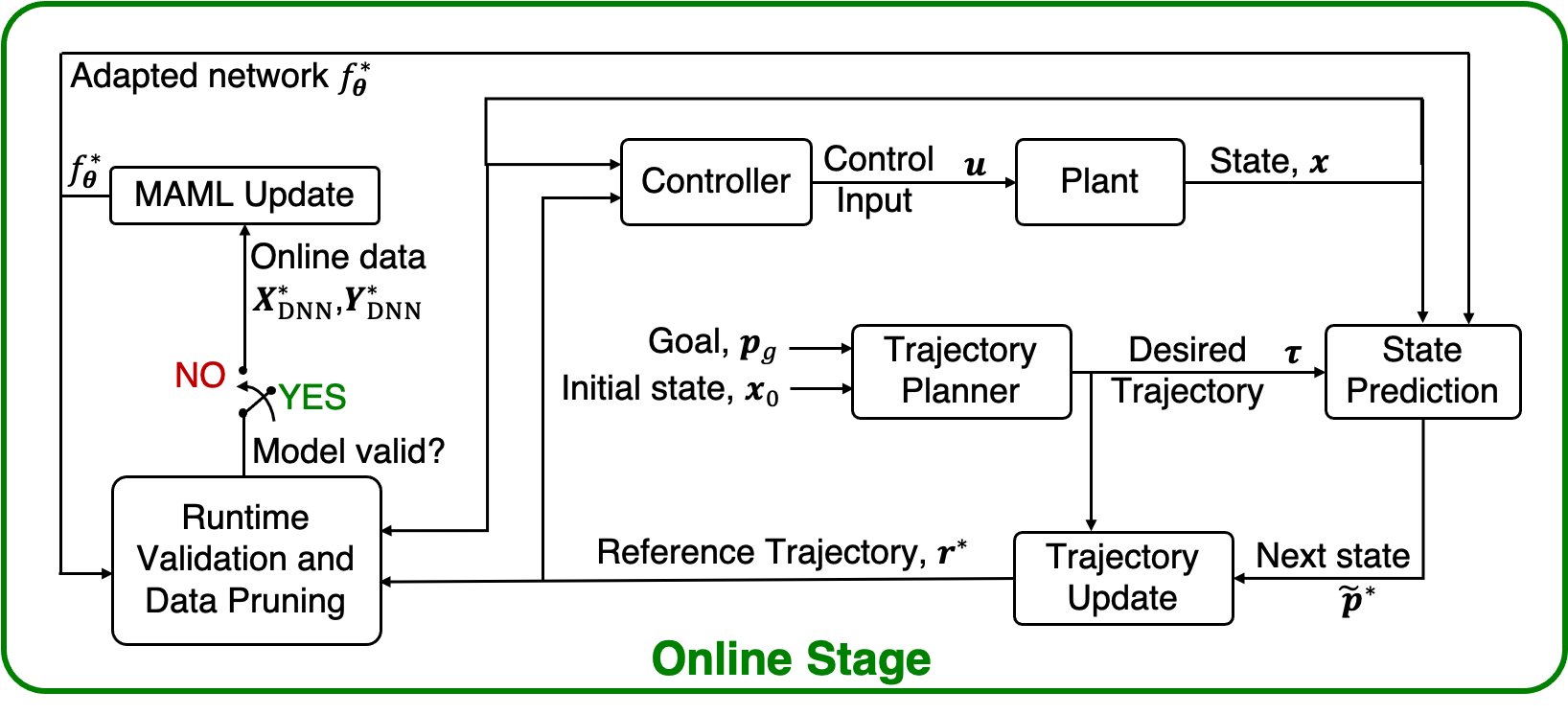}
	     \vspace{-15pt}
	    \caption{Architecture of the proposed runtime trajectory updating, runtime monitoring, and online learning approach.}
	    \label{fig:diagram_runtime}
	\end{figure}
	
	\subsubsection{Reference Trajectory Update using MAML}
	To explain the proposed reference trajectory update procedure, we use the pictorial representation in Fig.~\ref{fig:correction} as a guideline. By using the online training data shown by black crosses for $ K=5 $, the system fine-tunes its meta-trained model as explained in the previous section and predicts its next position:
	\vspace{-5pt}
	\begin{equation}
		\tilde{\bm{p}}^*(k+1) = \bm{f}_{\theta^*} \left(  \left[ \begin{array}{c}
		\bm{p}^*_{\tau}(k+1) \\ \bm{v}^*_{\tau}(k+1)		
		\end{array}\right]  -  \left[ \begin{array}{c}
		\bm{p}^*(k) \\ \bm{v}^*(k)		
		\end{array}\right]\right)   + \bm{p}^*(k)
		\label{eq:prediction}
	\end{equation}
	
	The next position prediction $ \tilde{\bm{p}}^*(k+1) $ (shown by the orange dot in Fig.~\ref{fig:correction}) is used to update the reference trajectory to compensate for the fault that the system is experiencing. By correcting the reference trajectory in the opposite side of the predicted deviation, the system is tricked to apply the necessary inputs to follow its desired path without a need for accessing the controller. The predicted deviation from the desired path is shown by orange dashed line in Fig.~\ref{fig:correction} and is calculated as follows:
	\begin{equation}
		\tilde{\bm{d}}(k+1) = \bar{\bm{p}}_\tau^*(k+1) - \tilde{\bm{p}}^*(k+1) 
		\label{eq:deviation}
	\end{equation}
	where $ \tilde{\bm{p}}^*(k+1) $ is the position prediction as calculated in \eqref{eq:prediction} and $ \bar{\bm{p}}_\tau^*(k+1) $ is the closest point on the trajectory $ \tau^* $ to the predicted position:
	\begin{equation}
		\bar{\bm{p}}_\tau^*(k+1) = \argmin_{\bm{p}_\tau^*(t), t \in \{0, \ldots, T(\tau_j)\}} \rVert \tilde{\bm{p}}^*(k+1) - \bm{p}^*_\tau(t)  \lVert
	\end{equation}
	
	\begin{figure}[b]
		\centering
		\subfigure[Trajectory update procedure.\label{fig:correction}]{\includegraphics[width=0.23\textwidth]{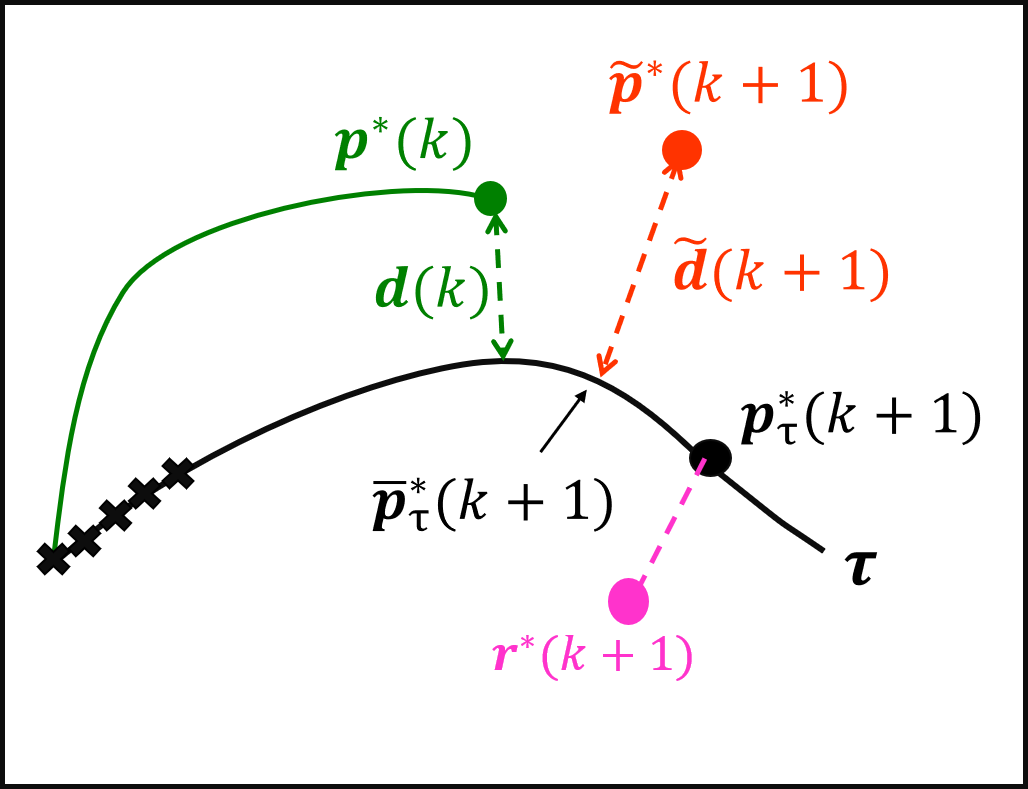}}
		\subfigure[Runtime validation and relearning procedure.\label{fig:relearning}]{\includegraphics[width=0.23\textwidth]{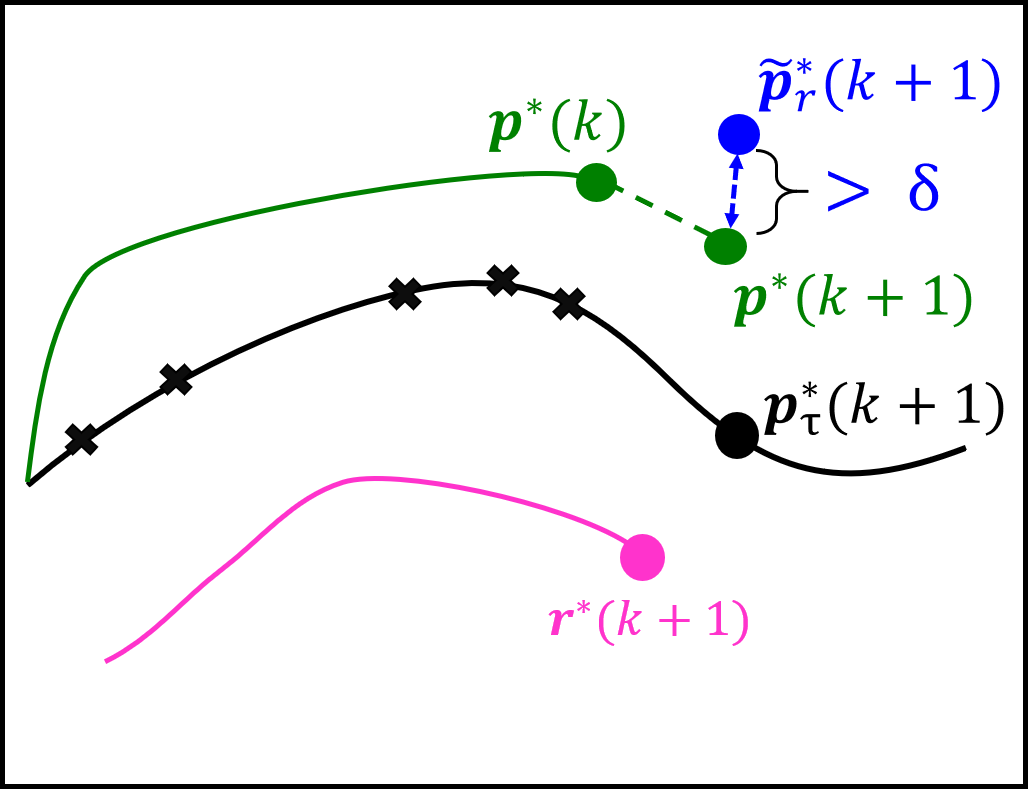}}
		\vspace{-5pt}
		\caption{Pictorial representations of the proposed trajectory update and online learning methods.}
	\end{figure}
	To compensate for the fault, the reference trajectory position (shown by magenta dot in Fig.~\ref{fig:correction}) is updated based on the original desired trajectory:
	\begin{equation}
	    \bm{r}^*(k+1) = \bm{p}_\tau^*(k+1) + \bm{c}(k+1) 
	    \label{eq:reference}
	\end{equation}
	where $ \bm{c}(k+1) $ is the trajectory update vector calculated based on the history of observed deviations and the predicted deviation using a PID-based rule:
	\begin{align}
	     \bm{c}(k+1) =\, & \kappa_p \tilde{\bm{d}}(k+1) + \kappa_d (\tilde{\bm{d}}(k+1) - \bm{d}(k)) \\ +\, &  \nonumber \kappa_i \left(\sum_{t=0}^{t=k} \bm{d}(t) + \tilde{\bm{d}}(k+1)\right)
	\end{align}
	where $ \bm{d}(k) $ (green dashed line in Fig.~\ref{fig:correction}) is the system's deviation from the desired path at time $ k $  and $ \kappa_p $, $ \kappa_d $ and $ \kappa_i $ are positive proportional, derivative and integral gains respectively. The reference velocity to the system is also updated accordingly:
	\begin{align}
        \bm{r}_v^*(k+1) = \frac{\bm{r}^*(k+1)-\bm{r}^*(k)}{\Delta k}
	\end{align}
	
	With this reference updating strategy, the faulty system converges to its desired path over time. After the updated reference is applied to the controller as an input, the learned model is validated by comparing the model predictions with the actual position of the system as described in the next section.

	\subsubsection{Runtime Model Validation}
	When the system is operating under a previously unseen fault, an initially adapted meta-trained model may not be able to make accurate predictions if the system's behavior has a lot of variations over time. To prevent inaccurate predictions and their potential negative effects at runtime, the model is constantly validated by comparing its prediction with the actual state. The proposed runtime model validation scheme is pictorially presented in Fig.~\ref{fig:relearning}. 	
	After updating the reference trajectory, the future position of the system $ \tilde{\bm{p}}_r^*(k+1) $ (shown by a blue point), while it's following the new reference trajectory $ \bm{r}^* $, is predicted using the fine-tuned model:
	 
	 \vspace{-15pt}
	\begin{equation}
		\tilde{\bm{p}}_r^*(k+1) = \bm{f}_{\theta^*} \left(  \left[ \begin{array}{c}
		\bm{r}^*(k+1) \\ \bm{r}_v^*(k+1)		
		\end{array}\right]  -  \left[ \begin{array}{c}
		\bm{p}^*(k) \\ \bm{v}^*(k)		
		\end{array}\right]\right)   + \bm{p}^*(k)
	\end{equation}
	If the predicted future position differs from the actual position one step later more than a given threshold (i.e., if the deviation shown by blue dashed line in Fig.~\ref{fig:relearning} exceeds a certain threshold), the learned model is invalidated:
	\begin{equation}
		 s(k+1) = \begin{cases}
		 0 & \text{if   }  \rVert \tilde{\bm{p}}_r^*(k+1) - \bm{p}^*(k+1) \lVert > \delta \\
		 1 & \text{otherwise}
		 \end{cases}
	\end{equation}
	where $ s(k+1) $ is a binary variable that denotes the validity of the learned model $ f_{\bm{\theta}}^* $ and $ \delta $ is a user-defined threshold for prediction deviation. If the model is invalidated ($ s = 0 $), online re-learning is triggered and the model is re-adapted 
	using new online data selected as explained next.
	
	\subsubsection{Data Pruning and Online Relearning} \label{sec:data_pruning}
	To quickly update the meta-learned model with the runtime data, the number of online training samples needs to be kept bounded. However, selecting data which are not representative of the system may cause poor learning performance, and lead to unnecessary relearning operations. To prevent this and relearn the model with the most representative data, we use k-means clustering \cite{kmeans-lloyd}.

	At time $ k+1 $ (for $ k > K $), the system collects data from the history of the relative reference positions and velocities:
	\begin{align}
	& \bm{X}_{*} = \left[ \begin{array}{c}
	\bm{r}^*(2:k+1) \\ \bm{r}_v^*(2:k+1)		
	\end{array}\right]  -  \left[ \begin{array}{c}
	\bm{p}^*(1:k) \\ \bm{v}^*(1:k)		
	\end{array}\right]  \nonumber \\
	& \bm{Y}_{*} = \bm{p}^*(2:k+1) - \bm{p}^*(1:k) 
	\end{align}
	where $ \bm{X}_{*} \in \mathbb{R}^{6\times k} $ is the history of potential inputs to the model learning and $ \bm{Y}_{*} \in \mathbb{R}^{3\times k} $ is the history of potential outputs. The history of inputs are clustered into $ K $ clusters and the centroid of each cluster $ j $ is obtained: $ \bm{C}_j \in \mathbb{R}^{6}, \forall j \in \{1, \ldots, K\} $. For online training, the closest point to each centroid in the input data history is selected as a training input:
	\vspace{-2pt}
	\begin{align}
		\bm{X}_\text{DNN}^*(j) = \argmin_{\bm{X}_{*}(n), n \in \{1, \ldots, k\}} \rVert \bm{X}_{*}(n) - \bm{C}_j \lVert  \text{  } \forall j \in \{1, \ldots, K\} \nonumber
	\end{align}
	\vspace{-2pt}
	The model is re-learned as explained in Section \ref{sec:online_maml} by using the pruned input data $ \bm{X}^*_\text{DNN} \in \mathbb{R}^{6 \times K} $ and the corresponding output data  $ \bm{Y}^*_\text{DNN} \in \mathbb{R}^{3 \times K} $. As pictorially shown in Fig.~\ref{fig:relearning}, with the proposed re-learning procedure the system picks new learning data from its history which are more representative and sparse than the initial data used. By using this data pruning and online learning approach, more diverse training data are chosen online, and redundant data are removed, leading to more efficient and effective online learning at runtime.
	
    It should be noted that the complexity of the k-means clustering increases with the number of data points to be clustered, which can potentially cause runtime problems for longer trajectories. This issue can easily be solved by limiting the data size by removing older data -- with the intuition that the significance of the previous data on the system's current behavior decreases over time.

\end{section}
  
\begin{section}{Simulations} \label{sec:simulation}
	
    \begin{figure}[b]
    \vspace{-15pt}
        \centering
        \subfigure[Path of a UAV with a fault in propeller 2 following a slalom path in between obstacles. \label{fig:sim1} ]{\includegraphics[width=0.48\textwidth]{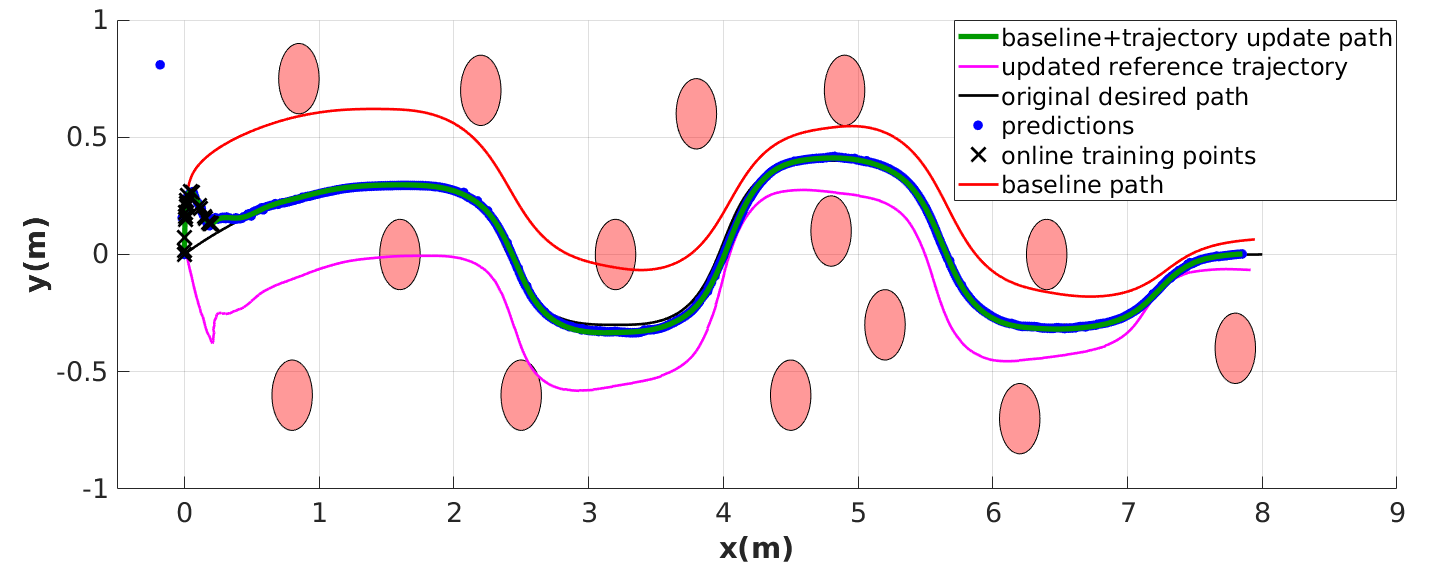}}
        \subfigure[Zoomed in version of the area marked with the box in Fig.~\ref{fig:sim1}. \label{fig:sim1_zoom} ]{\includegraphics[width=0.24\textwidth]{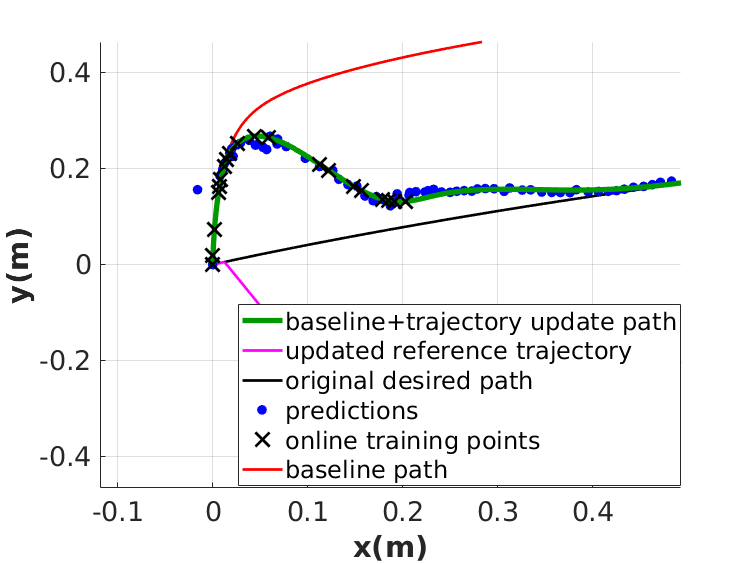}}
        \subfigure[Deviation over time with and without our approach for the case shown in Fig.~\ref{fig:sim1}. \label{fig:sim1_dev} ]{\includegraphics[width=0.22\textwidth]{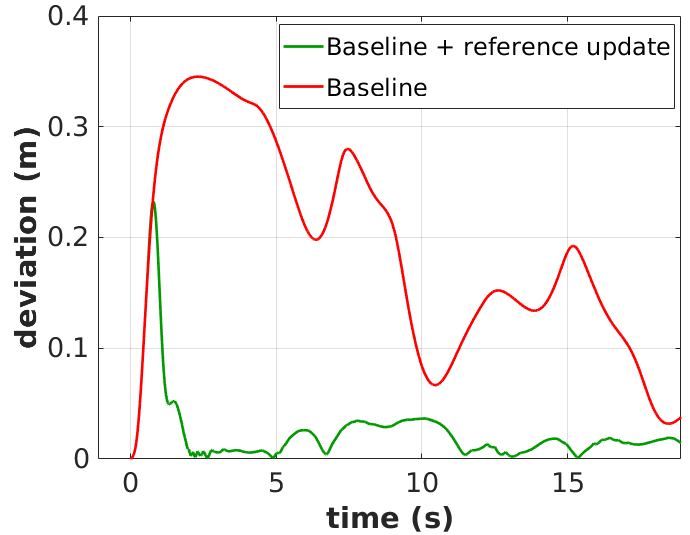}}
        \vspace{-5pt}
        \caption{Simulation results for UAV with $ \mathcal{F}_1^* $. \label{fig:simulations_fault1}}
    \end{figure}
	We validate the proposed meta-learning based reference trajectory adaptation, runtime monitoring, and re-learning approach with a quadrotor trajectory tracking case study. In this work, we use a \REV{12 dimensional} quadrotor UAV model with a baseline PID controller for position and attitude control which is designed for the nominal model (i.e., with no faults) \cite{KumarRAM2010}. The actuator fault is simulated by reduced thrust on various propellers. 
	During the offline stage, meta-training data are collected with a nominal UAV model and with UAV models under four different actuator faults given in Table~\ref{table:faults}. 
	
	To obtain enough training data, minimum-jerk trajectories \cite{Mellinger2011}, to four different goal position with different initial and final speed values are generated, and a faulty UAV is tasked to follow these trajectories. We utilized a Tensorflow Keras implementation of MAML \cite{finn2017maml} for training and adapting the meta-learning. We train a neural network with two hidden layers with 40 nodes using the offline training data.
	
	\begin{table}[h]
	\vspace{-5pt}
	\centering
	\caption{Fault types used during simulations.}
    \vspace{-4pt}
	\label{table:faults}
    \begin{tabular}{|c|c|}
    \hline
    \centering
    Training fault name & Fault type  \\ \hline
	$ \mathcal{F}_1 $ & 60\% of the commanded thrust on propeller 1  \\ \hline
	$ \mathcal{F}_2 $ & 80\% of the commanded thrust on propeller 1  \\ \hline
	$ \mathcal{F}_3 $ & 60\% of the commanded thrust on propeller 2 \\ \hline
	$ \mathcal{F}_4 $ &  80\% of the commanded thrust on propeller 2 \\ \hline  \hline
    Test fault name & Fault type  \\ \hline 
	$ \mathcal{F}^*_1 $ & 70\% of the commanded thrust on propeller 2  \\ \hline
	$ \mathcal{F}^*_2 $ & 60\% of the commanded thrust on propeller 4  \\ \hline
\end{tabular}
    \vspace{-5pt}
    \end{table}
    
    \begin{figure*}
        \centering
        \subfigure[UAV with fault 2 in the middle of its operation. \label{fig:sim2} ]{\includegraphics[width=0.39\textwidth]{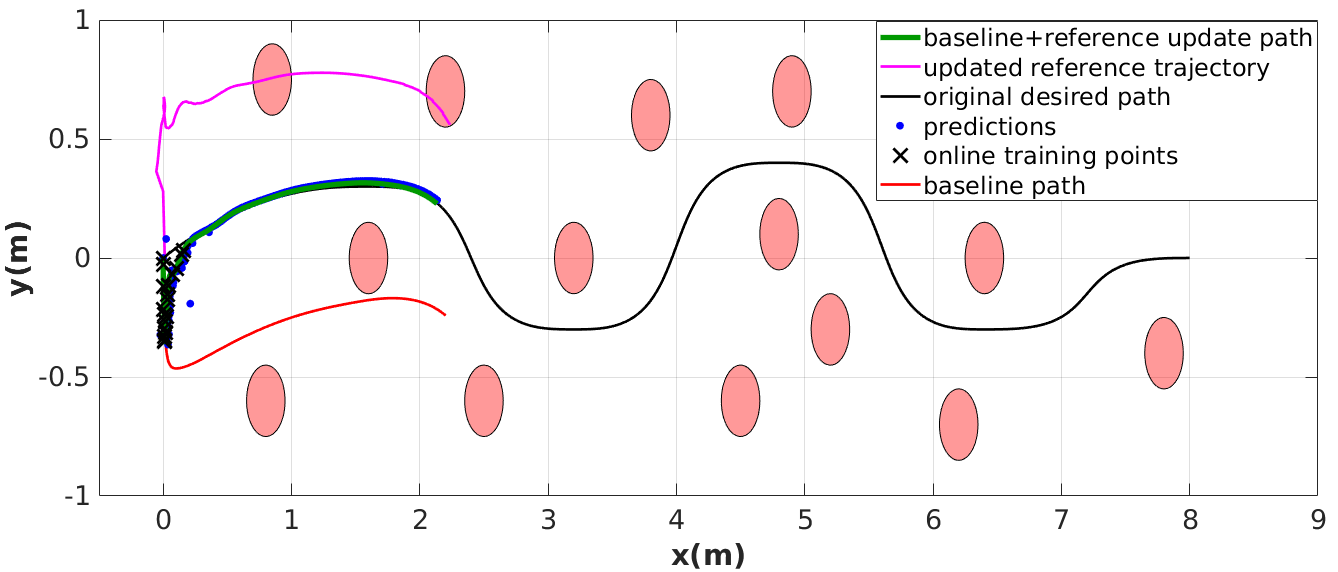}}
        \subfigure[The remaining part of the UAV operation highlighting different training points than in (a) for meta-learning. \label{fig:sim2_relearn} ]{\includegraphics[width=0.39\textwidth]{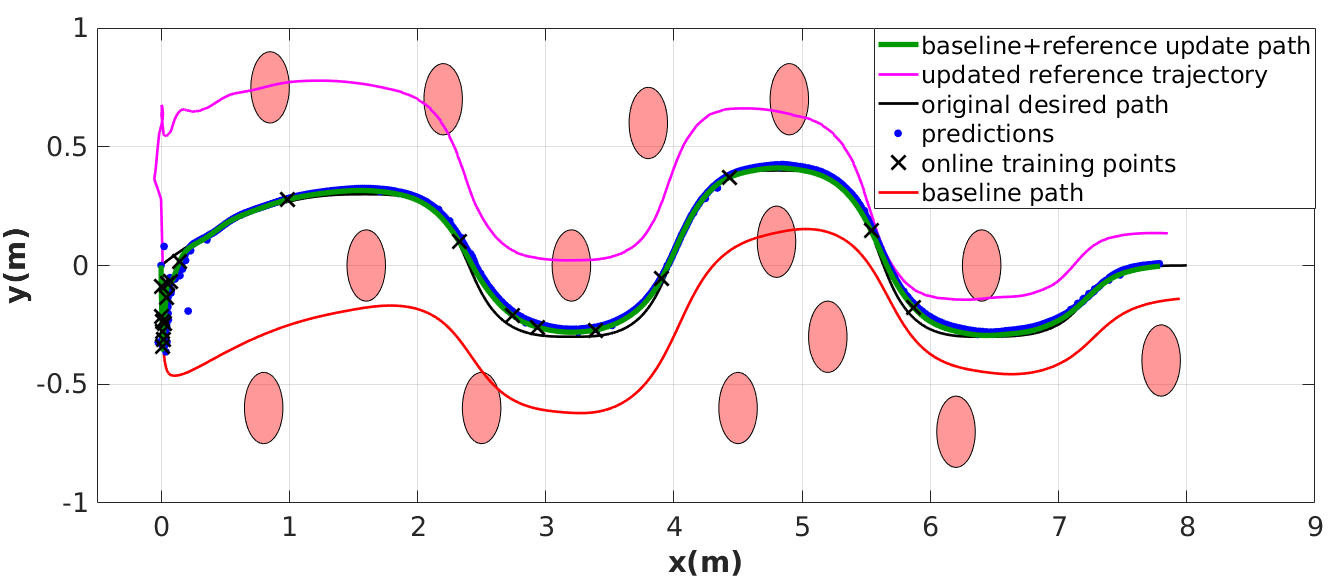}}
        \subfigure[Deviation over time. \label{fig:sim2_dev} ]{\includegraphics[width=0.19\textwidth]{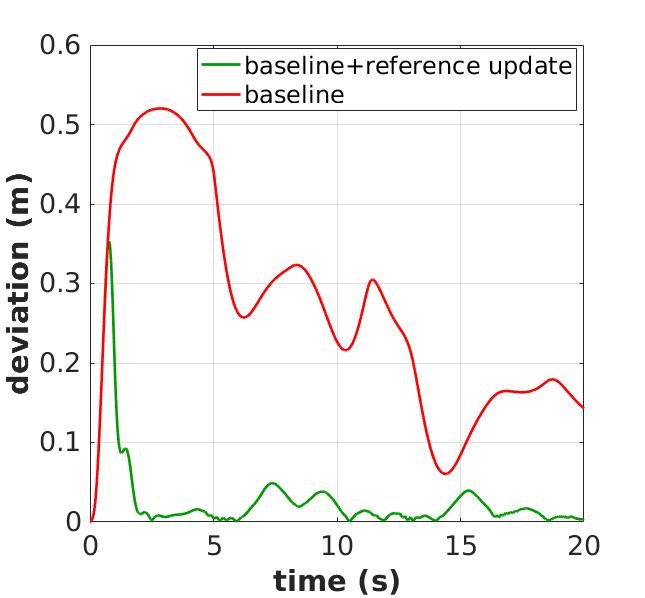}}
        \vspace{-5pt}
        \caption{Simulation results for UAV with fault $ \mathcal{F}_2^* $}
        \label{fig:simulations_fault2}
        \vspace{-15pt}
    \end{figure*}

	At runtime, the UAV is tasked to follow a path in an obstacle cluttered environment. In the case shown in Fig.~\ref{fig:simulations_fault1}, the system is tasked to move to a goal at $ \bm{p}_g = [8, 0, 1] $m from its initial position $ \bm{p}_0 = [0, 0, 1] $m following an obstacle-free minimum-jerk trajectory. We consider an actuator fault between $ \mathcal{F}_3 $ and $ \mathcal{F}_4 $ which is not used during the training $ \mathcal{F}^*_1 $: 70\% of the commanded thrust on propeller 2.

	In Fig.~\ref{fig:sim1}, the UAV follows the desired trajectory (black curve) moving between the obstacles (red semi-transparent circles). At the beginning of the operation, the UAV  collects $ K=20 $ training samples (black crosses) and updates its meta-trained neural network accordingly. Using the updated model, the UAV's next position is predicted and the reference trajectory inputted to the system is updated. The magenta curve shows the updated reference trajectory. After the trajectory update, the next position of the system tracking the updated reference is predicted (blue dots) and compared with the system's actual position (green curve) for validation and re-learning. 
	In Fig.~\ref{fig:sim1_zoom}, we show a zoomed-in version of the area marked with the box in Fig.~\ref{fig:simulations_fault1}(a) to show the data more clearly. Using our approach, the system deviates much less than the case where it follows the original trajectory and collides with obstacles (red curve). In Fig.~\ref{fig:sim1_dev}, the deviation from the desired path using our reference update approach on top of the baseline controller is compared with the case with the baseline controller. The average deviation from the desired trajectory is recorded as 18.17cm with the baseline controller and with our approach, it is reduced to 2.24cm. 
	The system performs relearning operations to adapt its model nine times at the beginning of the operation.

	In the case shown in Fig.~\ref{fig:simulations_fault2}, we consider a UAV with an actuator fault, which is outside of the training bounds $ \mathcal{F}^*_2 $: 60\% of the commanded thrust on propeller 4. Similar to the previous case, the vehicle collects training data at the beginning of its operation and adapts its meta-trained model accordingly. Using the adapted model, it makes predictions about its future states and updates the reference trajectory, as shown in Fig.~\ref{fig:sim2}. When the difference between the model predictions and the actual state exceeds the desired threshold $ \delta = 0.02 $m, the system prunes its history of observations, as explained in Section~\ref{sec:data_pruning}. In Fig.~\ref{fig:sim2_relearn},  the vehicle's complete path is shown by the green curve, and the black crosses represent the position data of the vehicle used for the last model re-learning. The system adapts its meta-trained model 122 times. The deviation from the desired path with proposed approach is compared to the baseline controller in Fig.~\ref{fig:sim2_dev}. The average deviation from the desired path is recorded as 2.61cm with our approach as opposed to 27.05cm with baseline controller without using any reference update.

\end{section}
\begin{section}{Experiments} \label{sec:experiments}

	We validated the proposed meta-learning based reference trajectory update approach with experiments on an Asctec Hummingbird quadrotor UAV implemented in ROS. We used a Vicon motion capture system to monitor the state of the quadrotor. The UAV was tasked to follow a minimum-jerk trajectory while experiencing an unknown fault. In these experiments, a fault was implemented as a bias on the commanded roll angle to the vehicle's attitude controller. 

	During the offline training, we generated a trajectory with three different average speed values: $ \mathcal{V} = \{ 0.25, 0.35, 0.45 \} $ m/s and five different faults with bias values: $ b \in  \{ -0.18, -0.12, -0.06, 0.06, 0.12 \} $rad. Using the data from these flights, a neural network was meta-trained. During the online stage, the UAV with an untrained fault was tasked to follow a trajectory with a speed value not included in $ \mathcal{V} $. At the beginning of its operation, the UAV adapted the meta-trained network using  $ K = 50 $ initial data points. The adapted network was used to make predictions and to update the reference trajectory according to \eqref{eq:reference}. During the experiments, we did not apply the trajectory update on the $ z $ axis as the faults considered did not cause $ z $ deviations.

    In Fig.~\ref{fig:exp_results1}, the results for a quadrotor under fault $ b = -0.14 $rad tasked to follow a desired trajectory (black curve) with an average speed of 0.5m/s which is outside of training \REV{distribution} is shown. As can be noted, with our approach, the quadrotor flew in close proximity of the original desired trajectory  (green curve). In contrast, with only the baseline controller, it deviated by a large amount (red curve), which is undesirable for both safety and liveness concerns.
    With the proposed approach, the average deviation was reduced to $ 8.09 $cm from $ 55.78 $cm. The comparison of the two behaviors is displayed also in Fig.\ref{fig:exp_results1}(b) by showing sequences of snapshots of the quadrotors moving inside our lab with the reference update approach on top of the baseline controller (green boxes) compared to the baseline controller (red boxes).
     Similarly, in Fig.~\ref{fig:exp_fault2},  the quadrotor with a bias $ b=-0.17 $rad followed the desired trajectory much more closely with an average deviation of $ 12.39 $cm as opposed to $ 79.73 $cm without using our proposed meta-learning based reference trajectory update. The corresponding sequence of snapshots of the quadrotors is given in Fig.~\ref{fig:exp_fault2_ss}. In the case demonstrated in Fig.~\ref{fig:exp_fault3}, the UAV experienced a bias $ b = -0.05 $rad and was tasked to move with the average speed of 0.3m/s. The average deviation was recorded as $ 10.23 $cm, while the average deviation with the baseline controller was $ 1.16 $m. The related overlapped sequence of snapshots of the quadrotors is presented in Fig.~\ref{fig:exp_fault3_ss}. Videos related to the simulations and experiments can be found in the supplementary media.
     

    	\begin{figure}[h]
    	\vspace{-5pt}
		\centering
		\subfigure[UAV paths. \label{fig:exp_fault1} ]{\includegraphics[width=0.24\textwidth]{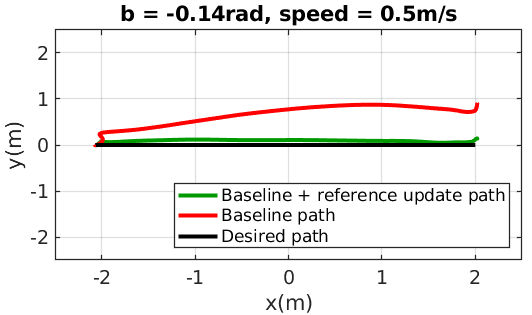}}
		\subfigure[Overlapped sequence of screenshots.\label{fig:exp_fault1_ss} ]{\includegraphics[width=0.21\textwidth]{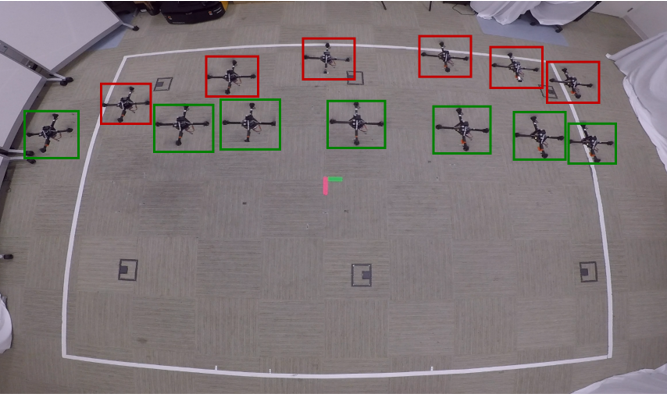}}
		\vspace{-5pt}
		\caption{Experiments results for the fault $ b = -0.14$rad with $ \bar{v} = 0.5 $m/s.\label{fig:exp_results1}}
		\vspace{-15pt}
	\end{figure}
	
	\begin{figure}[h]
	    \centering
	    \subfigure[UAV paths\label{fig:exp_fault2}]{\includegraphics[width=0.23\textwidth]{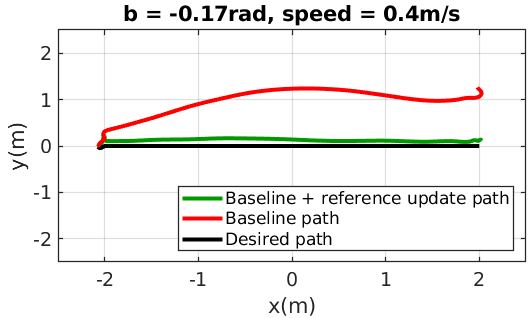}}
	    \subfigure[Overlapped sequence of screenshots.\label{fig:exp_fault2_ss}]{\includegraphics[width=0.21\textwidth]{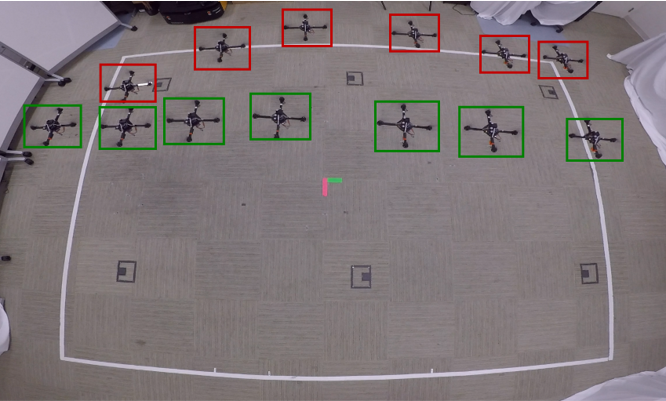}}
	    \vspace{-5pt}
	    \caption{Experiments results for the fault $ b = -0.17$rad with $ \bar{v} = 0.4 $m/s. \label{fig:exp_results2}}
	    \vspace{-15pt}
	\end{figure}
	
	\begin{figure}[h]
	    \centering
	    \subfigure[UAV paths.\label{fig:exp_fault3}]{\includegraphics[width=0.23\textwidth]{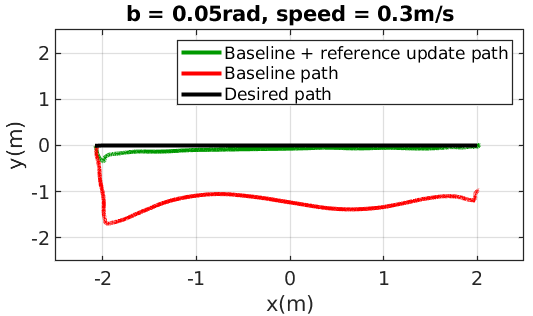}}
	    \subfigure[Overlapped sequence of screenshots.\label{fig:exp_fault3_ss}]{\includegraphics[width=0.21\textwidth]{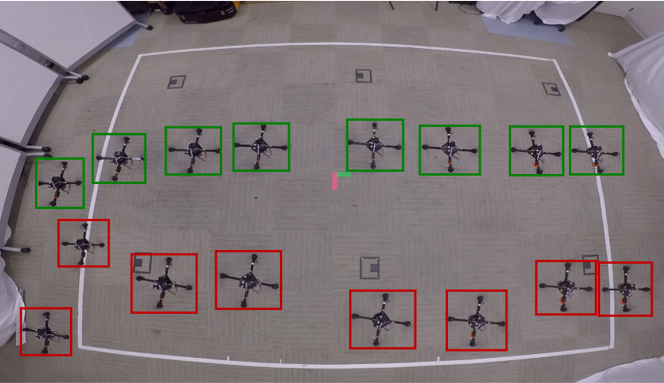}}
	    \vspace{-5pt}
	    \caption{Experiments results for the fault $ b = 0.05$rad with $ \bar{v} = 0.3 $m/s.\label{fig:exp_results3}}
	    \vspace{-5pt}
	\end{figure}

\end{section}

\vspace{-5pt}
\begin{section}{Conclusions and Future Work} \label{sec:conclusion} 
    \REV{In this work, we have presented an adaptive trajectory tracking approach for UAVs moving under unforeseen actuator faults. We have leveraged meta-learning to train easily adaptable models for prediction and reference trajectory updates.} The proposed reference trajectory update method makes the system with an actuator fault follow the desired trajectory better without needing access to the control inputs. With the proposed runtime validation and data pruning scheme, the updated meta-learned model is continuously monitored and relearned with a small number of representative data at runtime when necessary.
	
	This work opens an exciting path towards using meta-learning approaches for UAV fault rejection.
	\REV{We are currently working on extending this work by developing a meta-learning-based safety monitor to detect and avoid unsafe situations over a given future time horizon. As part of future work, the reference trajectory update method can be improved by leveraging gain scheduling or learning techniques. Additionally, learning approaches can also be used to monitor the validity of the runtime updated network.}
	
	
\end{section}

\section*{Acknowledgments} 
\vspace{-5pt}
This work is based on research sponsored by DARPA under Contract No. FA8750-18-C-0090.

\bibliographystyle{IEEEtran}
\bstctlcite{IEEEexample:BSTcontrol}
\bibliography{ms.bib}

\end{document}